# Finite Hilbert Transform in Weighted L² Spaces[1]

## Jiangsheng You

Cubic Imaging LLC, 18 Windemere Dr., Andover, MA 01810
jshyou@gmail.com

**Abstract**

Several new properties of $\int_{-1}^{1}[f(t)\cosh(\mu(s-t))/(s-t)]dt$ in the weighted $L^2$ spaces are obtained. If $\mu = 0$, two Plancherel-like equalities and the isotropic properties are derived. For $\mu \in R^1$, a coerciveness inequality is derived and two iterative sequences are constructed to find the inversion. The proposed iterative sequences are applicable to the case of pure imaginary constant $\mu = i\eta$ with $|\eta| < \pi/4$. For $\mu = 0, 3.0$, we present the computer simulation results by using the Chebyshev series representation of finite Hilbert transform.

## I.   Mathematical preliminary and application background

The investigation of the finite Hilbert transform (FHT) has a long history in the literature, for example, [1-7, 9, 11, 13, 14, 17, 20, 22] and the references therein. The singular integral operator theory is the classical method in the study of the FHT. In particular, the Poincaré-Bertrand formula plays the most important role in these existing works. Besides the pure mathematical interest, there have been several practical applications of using the FHT, for example, from fluid mechanic [10, 23] to medical imaging in [8, 15-16, 21, 24-29].

Let $f(t)$ be a function of $R^1$ and $F(s)$ denote its Hilbert transform. They are related by

$$F(s) = \frac{1}{\pi} \int_{-\infty}^{\infty} \frac{f(t)}{s-t} dt \text{ and } f(t) = \frac{1}{\pi} \int_{-\infty}^{\infty} \frac{F(s)}{s-t} ds. \qquad (1.1)$$

Throughout this paper, the integral at the singular point should be understood in the sense of Cauchy principal value. Let $L^1(R^1)$ and $L^2(R^1)$ be the Banach spaces of integrable and square-integrable functions in $R^1$, respectively. If $f(t) \in L^1(R^1) \cap L^2(R^1)$, equations of (1.1) hold almost everywhere in $R^1$, and satisfy the following Plancherel equality:

$$\int_{-\infty}^{\infty} |F(s)|^2 \, ds = \int_{-\infty}^{\infty} |f(t)|^2 \, dt. \qquad (1.2)$$

Notice that the integrals in equations (1.1) and (1.2) are over the whole real axis.

The focus of this paper will be the Hilbert transform over the open unit interval $I = (-1, 1)$. Denote by $C_0^\infty(I)$ the linear space of infinitely differentiable functions with support in $I$. In contrast to (1.1), in the remaining of this paper we mean the FHT in $(-1, 1)$ by

$$F(s) = \frac{1}{\pi} \int_{-1}^{1} \frac{f(t)}{s-t} dt, \qquad (1.3)$$

here both $f(t)$ and $F(s)$ are defined in $I$. For $1 < p < \infty$, the Banach space $L^p(I)$ is defined as

$$L^p(I) = \{f(t) : \int_{-1}^{1} |f(t)|^p \, dt < \infty\}. \qquad (1.4)$$

---

From [14, 20, 22], the FHT $F(s)$ is continuous in $L^p(I)$, $1<p<\infty$. For $f \in L^p(I)$ and $g \in L^q(I)$ with $q=p/(p-1)$, in [11, 13, 14, 22], the Poincaré-Bertrand formula can be expressed as

$$\int_{-1}^{1}[\int_{-1}^{1}\frac{f(u)}{\pi(s-u)}du]\frac{g(s)}{\pi(s-t)}ds + \int_{-1}^{1}[\int_{-1}^{1}\frac{g(s)}{\pi(u-s)}ds]\frac{f(u)}{\pi(u-t)}du$$
$$= f(t)g(t) - [\int_{-1}^{1}\frac{f(u)}{\pi(t-u)}du][\int_{-1}^{1}\frac{g(s)}{\pi(t-s)}ds]. \tag{1.5}$$

We mention that (1.5) appeared in several different formats in the literature [14, 22]. Hereafter we define weight function $w(t)=\sqrt{1-t^2}$. We cite two equalities in [22] as follows

$$\frac{1}{\pi}\int_{-1}^{1}\frac{w(t)}{s-t}dt = -s, \qquad \frac{1}{\pi}\int_{-1}^{1}\frac{1}{s-t}\frac{1}{w(t)}dt = 0. \tag{1.6}$$

For $f \in C_0^\infty(I)$, replacing $g(t)$ by $w(t)$ and $1/w(t)$ in (1.5) correspondingly, we deduce two well-known inversion formulae

$$f(t) = \frac{1}{w(t)}[\int_{-1}^{1}\frac{F(s)}{\pi(s-t)}w(s)ds + \frac{1}{\pi}\int_{-1}^{1}f(u)du], \tag{1.7}$$

$$f(t) = w(t)\int_{-1}^{1}\frac{F(s)}{\pi(s-t)}\frac{1}{w(s)}ds. \tag{1.8}$$

It was proved in [17] that if $f \in L^2(I)$, $\int_{-1}^{1}f(u)du$ can be calculated by the following limit:

$$\int_{-1}^{1}f(u)du = -\pi\lim_{\varepsilon\to 0}\{\int_{\varepsilon-1}^{1-\varepsilon}\frac{\hat{f}(t)}{w(t)}dt / \int_{\varepsilon-1}^{1-\varepsilon}\frac{dt}{1-t^2}\}, \tag{1.9}$$

here $\hat{f}(t) = \int_{-1}^{1}[F(s)/(s-t)][w(s)/w(t)]/\pi ds$. Thus the combination of (1.7) and (1.9) becomes an explicit inversion formula for $f \in L^2(I)$. We refer to [14, 22] for more inversion formulae and functional properties in $L^p(I)$. The range condition in $L^p(I)$ for $p>2$ has been well defined. For reference we state the range condition in Lemma 1.

**Lemma 1**. *If $f \in C_0^\infty(I)$, then $F(s)$ satisfies the following equation*:

$$\int_{-1}^{1}\frac{F(s)}{w(s)}ds \equiv 0. \tag{1.10}$$

**Proof**. The proof can be found in [14, 22]. □

Next we give an introduction to the method that reduces the image reconstruction in single photo emission computed tomography (SPECT) to inverting a cosh-weighted Hilbert transform. Let $\mu \in R^1$ and $g(x,y)$ be a function of $R^2$. The exponential Radon transform (ERT) is defined as

$$[R_\mu g](\theta, s) = \int_{-\infty}^{\infty}g(s\vec{\theta}+t\vec{\theta}^\perp)e^{\mu t}dt, \tag{1.11}$$

where $\vec{\theta}=(\cos\theta, \sin\theta)$ and $\vec{\theta}^\perp=(-\sin\theta, \cos\theta)$. In SPECT, $[R_\mu g](\theta,s)$ can be obtained from the acquired data in $[-0.5\pi, -0.5\pi]\times R^1$. Let $\vec{r}=(x,y)$. Simple calculations such as [21, Section 3] lead to the following relation

$$\frac{1}{2\pi}\int_{-\pi/2}^{\pi/2}e^{-\mu\vec{r}\cdot\vec{\theta}^\perp}\frac{\partial[R_\mu g](\theta, s)}{\partial s}\Big|_{s=\vec{r}\cdot\vec{\theta}}d\theta = \int_{-\infty}^{\infty}\frac{\cosh(\mu\tau)g(x-\tau, y)}{-\pi\tau}d\tau. \tag{1.12}$$

Then the reconstruction of $g(x,y)$ from $R_\mu g$ becomes equivalent to inverting the right hand side of (1.12) which is a cosh-weighted Hilbert transform defined by

$$F_\mu(s) = \frac{1}{\pi} \int_{-1}^{1} \frac{\cosh(\mu t)}{t} f(s-t)dt. \tag{1.13}$$

In SPECT, the constant $\mu$ takes non-negative values. As discussed in [15, 26], the imaging problem in reflection ultrasound tomography and seismic imaging can be mathematically modeled with a special case with a pure imaginary constant $\mu = i\eta$. For computed tomography (CT), which is equivalent to a special case with $\mu = 0$ in (1.12), the inversion formulae (1.7) and (1.8) can be used for the image reconstruction as shown in many works published in the journals: *IEEE Transactions on Medical imaging*, *Inverse problems* and *Physics in Medicine and Biology*.

Now we give a brief overview on the inversion of (1.13). The first inversion result was obtained in [21] in which $F_\mu(s)$ is required to be known on a larger interval of [–2, 2] in order to completely reconstruct $f(t)$ by using a polynomial series expansion. An inversion method using (1.7) and the power series expansion was numerically implemented for $\mu \leq 8$ in [16]. In [8], we used (1.8) to construct a different numerical inversion algorithm to reconstruct simulated SPECT data under various data acquisition settings. Nonetheless, no mathematical proof has been obtained on the inversion of (1.13) in a linear functional space when the author started this paper in 2011. In 2013, some remarkable results appeared in [2] on the range condition and explicit inversion formulas of (1.3) in $L^p(I)$. In this paper, we will prove the existence and uniqueness of the inversion of (1.13) in the weighted $L^2$ spaces.

The main results in this paper are as follows. For $\mu = 0$, we derive two Plancherel-like equalities and the isotropic properties of the FHT in the weighted $L^2$ spaces. The results in [17] for the conventional $L^2$ space are thus further refined. We prove the coerciveness of (1.13) to establish the existence and uniqueness of the inverse cosh-weighted FHT for $\mu \in R^1$. Then we construct two iterative sequences to find the solution of (1.13) in the weighted $L^2$ spaces. This confirms that the numerical solutions from [8, 16, 21] are unique. The iterative sequences are also applicable to $\mu = i\eta$ for $|\eta| < \pi/4$, and this provides a different estimate to ensure the stable inversion of cos-weighted FHT in [15, 26]. We also use the discrete cosine and sine transforms to explicitly express the FHT and its inversion in the numerical realization. Computer simulation results are included to show the feasibility of the proposed inversion methods for practical applications.

## II.  Plancherel-like Equalities in Weighted L² Spaces

This section investigates the Plancherel-like equalities in the weighted $L^2(I)$ spaces. Two linear weighted spaces are defined as follows

$$L^2_m(I) = \{f(t): \int_{-1}^{1} |f(t)|^2 w(t)dt < \infty\}, \tag{2.1}$$

$$L^2_d(I) = \{f(t): \int_{-1}^{1} \frac{|f(t)|^2}{w(t)} dt < \infty\}. \tag{2.2}$$

Subscriptions $m$ and $d$ stand for the multiplication and division of the weight $w(t)$ in the integrals, respectively. We mention that $L^2_d(I) \subset L^2(I) \subset L^2_m(I)$ and $C_0^\infty(I)$ is dense in these three linear spaces. Notice $\|1/w(t)\|_{L^2_m} = 1$ and $\|1\|_{L^2_d} = 1$, but $1/w(t) \notin L^2(I)$. Weighted spaces $L^2_m(I)$ and $L^2_d(I)$ are actually Hilbert spaces with the following inner products

$$<f,g>_m = \frac{1}{\pi}\int_{-1}^{1} f(t)g^*(t)w(t)dt, \quad <f,g>_d = \frac{1}{\pi}\int_{-1}^{1} \frac{f(t)g^*(t)}{w(t)}dt. \tag{2.3}$$

Hereafter, the superscript star stands for the complex conjugate. Define two closed subspaces $Ł_m^2(I)$ and $Ł_d^2(I)$ as

$$Ł_m^2(I) = \{f(t) \in L_m^2(I) : <f, \frac{1}{w(t)}>_m \equiv 0\}, \quad Ł_d^2(I) = \{f(t) \in L_d^2(I) : <f, 1>_d \equiv 0\}. \tag{2.4}$$

Because $L_m^2(I)$ and $L_d^2(I)$ are inner product spaces, it follows that $L_m^2(I) = Ł_m^2(I) \oplus \{C/w(t)\}$ and $L_d^2(I) = Ł_d^2(I) \oplus \{C\}$, where $C \in R^1$. The range condition (1.10) is equivalent to $F(s) \in Ł_d^2(I)$, and $Ł_m^2(I)$ is the subspace of $L_m^2(I)$ with $\int_{-1}^{1} f(t)dt = 0$. This section we obtain several new properties of the FHT (1.3) in the weighted spaces $L_m^2(I)$ and $L_d^2(I)$.

**Lemma 2**. If $f(t) \in L_m^2(I)$, then

$$\int_{-1}^{1} |F(s)|^2 w(s)ds = \int_{-1}^{1} |f(t)|^2 w(t)dt - \frac{1}{\pi} \int_{-1}^{1} f^*(s)ds \int_{-1}^{1} f(u)du \leq \int_{-1}^{1} |f(t)|^2 w(t)dt. \tag{2.5}$$

**Proof**. First, we prove (2.5) for $f \in C_0^\infty(I)$. By taking $g = f^*$ in (1.5), multiplying $w(t)$ to each side of (1.5), integrating over $I$ and using the first equality of (1.6), we obtain

$$\int_{-1}^{1} |F(t)|^2 w(t)dt - \int_{-1}^{1} |f(t)|^2 w(t)dt$$

$$= \int_{-1}^{1} w(t)\{\int_{-1}^{1} \frac{f(u)}{\pi(t-u)}du \int_{-1}^{1} \frac{f^*(u)}{\pi(t-u)}du + f(t)f^*(t)]\}dt$$

$$= \int_{-1}^{1} w(t)dt\{\int_{-1}^{1} [\frac{f^*(s)}{\pi(t-s)} \int_{-1}^{1} \frac{f(u)}{\pi(s-u)}du + \frac{f(s)}{\pi(t-s)} \int_{-1}^{1} \frac{f^*(u)}{\pi(s-u)}du]ds\}$$

$$= -\int_{-1}^{1} [f^*(s) \int_{-1}^{1} \frac{f(u)}{\pi(s-u)}du + f(s) \int_{-1}^{1} \frac{f^*(u)}{\pi(s-u)}du]sds$$

$$= -\int_{-1}^{1} f^*(s) \int_{-1}^{1} \frac{f(u)s}{\pi(s-u)}duds - \int_{-1}^{1} f(s) \int_{-1}^{1} \frac{f^*(u)s}{\pi(s-u)}duds$$

$$= -\int_{-1}^{1} f^*(s)[\int_{-1}^{1} \frac{f(u)}{\pi}du + \int_{-1}^{1} \frac{f(u)u}{\pi(s-u)}du]ds - \int_{-1}^{1} f(s) \int_{-1}^{1} \frac{f^*(u)s}{\pi(s-u)}duds$$

$$= -\frac{1}{\pi} \int_{-1}^{1} f^*(s)ds \int_{-1}^{1} f(u)du + \int_{-1}^{1} f(u)u \int_{-1}^{1} \frac{f^*(s)}{\pi(u-s)}dsdu - \int_{-1}^{1} f(s)s \int_{-1}^{1} \frac{f^*(u)}{\pi(s-u)}duds$$

$$= -\frac{1}{\pi} \int_{-1}^{1} f^*(s)ds \int_{-1}^{1} f(u)du \leq 0. \tag{2.6}$$

Since $C_0^\infty(I)$ is a dense subspace in $L_m^2(I)$, equation (2.5) holds for $f(t) \in L_m^2(I)$. □

**Corollary 1**. If $f(t) \in Ł_m^2(I)$, then

$$\int_{-1}^{1} |F(s)|^2 w(s)ds = \int_{-1}^{1} |f(t)|^2 w(t)dt. \tag{2.7}$$

**Lemma 3**. If $f(t) \in L_d^2(I)$, then $F(s)$ satisfies (1.10) and

$$\int_{-1}^{1} \frac{|F(s)|^2}{w(s)}ds = \int_{-1}^{1} \frac{|f(t)|^2}{w(t)}dt. \tag{2.8}$$

**Proof**. For $f \in C_0^\infty(I)$, Lemma 1 yields that $F(s)$ meets (1.10). By taking $g = f^*$ in (1.5), multiplying $1/w(t)$ to each side of (1.5), integrating over $I$ and using the first equality of (1.6), we notice that the left hand side of (1.5) vanishes and then

$$\int_{-1}^{1}\frac{|F(s)|^{2}}{w(s)}ds=\int_{-1}^{1}\frac{|f(t)|^{2}}{w(t)}dt. \qquad (2.9)$$

Since $C_{0}^{\infty}(I)$ is a dense subspace of $L_{d}^{2}(I)$, thus (2.8) holds for $f(t)\in L_{d}^{2}(I)$. ▫

It is known from [14, 22] that (1.10) is sufficient for the existence of a unique $f\in L^{p}(I)$ for $p>2$. Theorem 1 presents more isotropic properties of the FHT $F(s)$ in $L_{d}^{2}(I)$ and $L_{m}^{2}(I)$. The proof of theorem 1 is through Lemmas 2 and 3, which is different from the arguments used in [17].

**Theorem 1**. *The FHT defined by (1.3) is isotropic from $L_{d}^{2}(I)$ to $Ł_{d}^{2}(I)$, and from $Ł_{m}^{2}(I)$ to $L_{m}^{2}(I)$. If $F(s)\in Ł_{d}^{2}(I)$, the inverse FHT $f(t)$ is determined by (1.8). If $F(s)\in L_{m}^{2}(I)$, the inverse FHT $f(t)$ in $Ł_{m}^{2}(I)$ is defined by*

$$f(t)=\frac{1}{w(t)}\int_{-1}^{1}\frac{F(s)}{\pi(s-t)}w(s)ds. \qquad (2.10)$$

**Proof**. We prove the first half of the theorem for $F(s)\in Ł_{d}^{2}(I)$. If $f(t)\in L_{d}^{2}(I)$, by Lemma 3, $F(s)\in Ł_{d}^{2}(I)$ and $\|F(s)\|_{L_{d}^{2}}=\|f(t)\|_{L_{d}^{2}}$. If $F(s)\in Ł_{d}^{2}(I)$, then $F(s)/w(s)\in L_{m}^{2}(I)$. Notice that $f(t)/w(t)$ is the FHT of $F(s)/w(s)$ in (1.8), $f(t)/w(t)$ meets (1.10) and $f(t)/w(t)\in Ł_{m}^{2}(I)$. By (2.7) in Corollary 1, hence $f(t)\in L_{d}^{2}(I)$ and $\|f(t)\|_{L_{d}^{2}}=\|F(s)\|_{L_{d}^{2}}$.

Next we prove the second half of the theorem for $F(s)\in L_{m}^{2}(I)$. If $f(t)\in Ł_{m}^{2}(I)$, by Corollary 1, $F(s)\in L_{m}^{2}(I)$ and $\|F(s)\|_{L_{m}^{2}}=\|f(t)\|_{L_{m}^{2}}$. If $F(s)\in L_{m}^{2}(I)$, then $\hat{F}(s)=F(s)w(s)\in L_{d}^{2}(I)$. Let $\hat{f}(t)=f(t)w(t)$, equation (2.10) becomes

$$\hat{f}(t)=\int_{-1}^{1}\frac{\hat{F}(s)}{\pi(s-t)}ds. \qquad (2.11)$$

Lemma 3 implies that $\hat{f}(t)\in L_{d}^{2}(I)$ and $\|f(t)\|_{L_{m}^{2}}=\|\hat{f}(t)\|_{L_{d}^{2}}=\|\hat{F}(s)\|_{L_{d}^{2}}=\|F(s)\|_{L_{m}^{2}}$. Because $\hat{f}(t)$ is the FHT of $\hat{F}(s)$, it meets (1.10) and $f(t)\in Ł_{m}^{2}(I)$. The theorem is thus proven. ▫

In Section I, we state that for $f\in L^{p}(I)$ and $g\in L^{q}(I)$ with $q=p/(p-1)$, the Poincaré-Bertrand formula is expressed in (1.5). As a direct consequence of Theorem 1 and the definitions of $L_{d}^{2}(I)$ and $L_{m}^{2}(I)$, (1.5) also holds for a pair of functions in $L_{d}^{2}(I)$ and $L_{m}^{2}(I)$, respectively.

**Corollary 2**. *Poincaré-Bertrand formula (1.5) holds for a pair of $f(t)\in L_{d}^{2}(I)$ and $g(t)\in L_{m}^{2}(I)$, or a pair of $f(t)\in L_{m}^{2}(I)$ and $g(t)\in L_{d}^{2}(I)$.*

**Proof**. For a fixed $g(t)\in L_{m}^{2}(I)$, the FHT of $g(t)$ belongs to $L_{m}^{2}(I)$ and equation (1.5) holds for any $f(t)\in C_{0}^{\infty}(I)$. If $f(t)\in L_{d}^{2}(I)$, $f(t)g(t)\in L^{1}(I)$ and (1.3) exists for functions in $L^{1}(I)$ from the note in [4], thus all the integrals on both sides of (1.5) are well defined. It follows that (1.5) holds for the pair of $f(t)\in L_{d}^{2}(I)$ and $g(t)\in L_{m}^{2}(I)$ by passing the limit from functions in $C_{0}^{\infty}(I)$ to $f(t)$ in $L_{d}^{2}(I)$. Similarly, equality (1.5) holds for the pair of $f(t)\in L_{m}^{2}(I)$ and $g(t)\in L_{d}^{2}(I)$. ▫

**Remark 1**. From [17], the FHT (1.3) is injective but not surjective from $L^{2}(I)$ to $L^{2}(I)$ since the range of the FHT in $L^{2}(I)$ is only a dense subspace of $L^{2}(I)$. For example, if $F(s)=1$, the solution to (1.3) is $-t/w(t)$ which does not belong to $L^{2}(I)$, thus $F(s)=1$ is not in the range of the FHT in $L^2(I)$. However, Lemmas 2 and 3 indicate that the FHT (1.3) is isotopic from $Ł^2_m(I)$ to $L^2_m(I)$ and $L^2_d(I)$ to $Ł^2_d(I)$, respectively.

## III. Cosh-Weighted FHT in Weighted L² Spaces

For a real constant $\mu \in R^1$ and a function $f(t)$ in $I$, we rewrite (1.13) as

$$F_\mu(s) = \frac{1}{\pi}\int_{-1}^{1}\frac{\cosh[\mu(s-t)]}{s-t}f(t)dt. \quad (3.1)$$

Equation (3.1) was first studied in [21] for the inverse exponential Radon transform (ERT) with 180º data, in which one inversion formula using Taylor series expansion was derived if $F_\mu(s)$ is known in the interval $(-2, 2)$. If $\int_{-1}^{1}\cosh(\mu t)f(t)dt$ is known, an inversion method using (1.7) and the power series expansion was numerically implemented for $\mu \leq 8$ in [16] but that procedure is not stable. Shortly after, without using $\int_{-1}^{1}\cosh(\mu t)f(t)dt$, we studied a simpler method to invert (3.1) via (1.8) in [8]. We mention that the existence and uniqueness of the inversion of (3.1) for all $\mu$ is still open in [8, 16, 21]. Assume $\mu \in R^1$, we will prove that $f(t)$ can be uniquely solved from (3.1) in the weighted spaces $L^2_m(I)$ and $L^2_d(I)$. More elegant results appeared in [2] on the inversion of (3.1). The author likes to emphasizes two significant results: one is the null space function (3.11) of [2] and the other is inversion formula (5.8-5.9) of [2].

**Theorem 2**. *Let $f(t) \in L^2_d(I)$. There exists a constant $a > 0$, depending only on $\mu \in R^1$, such that*

$$\|F_\mu(s)\|_{L^2_d} \geq a\|f(t)\|_{L^2_d}. \quad (3.2)$$

*Furthermore, the solution of (3.1) is unique in $L^2_d(I)$.*

**Proof**. With $\cosh[\mu(s-t)] = \cosh(\mu s)\cosh(\mu t) - \sinh(\mu s)\sinh(\mu t)$, we rewrite (3.1) as

$$\frac{F_\mu(s)}{\cosh(\mu s)} = \int_{-1}^{1}\frac{\cosh(\mu t)f(t)}{\pi(s-t)}dt - \tanh(\mu s)\int_{-1}^{1}\frac{\cosh(\mu t)f(t)}{\pi(s-t)}\tanh(\mu t)dt. \quad (3.3)$$

Denote by $F_{\mu,1}(s)$ and $F_{\mu,2}(s)$ two terms of right hand side of (3.3) as follows

$$F_{\mu,1}(s) = \int_{-1}^{1}\frac{\cosh(\mu t)f(t)}{\pi(s-t)}dt, \quad F_{\mu,2}(s) = \tanh(\mu s)\int_{-1}^{1}\frac{\cosh(\mu t)f(t)}{\pi(s-t)}\tanh(\mu t)dt. \quad (3.4)$$

Notice that $\cosh(\mu t)f(t) \in L^2_d(I)$ and $|\tanh(\mu s)| \leq |\tanh(\mu)| < 1$ for $s \in I$. By Lemma 3, we have

$$\|F_{\mu,1}(s)\|^2_{L^2_d} = \|\cosh(\mu t)f(t)\|^2_{L^2_d}, \quad (3.5)$$

$$\|F_{\mu,2}(s)\|^2_{L^2_d} \leq \tanh^2(\mu)\|\sinh(\mu t)f(t)\|^2_{L^2_d} \leq \tanh^4(\mu)\|\cosh(\mu t)f(t)\|^2_{L^2_d}. \quad (3.6)$$

By the triangular inequality, we obtain

$$\left\|\frac{F_\mu(s)}{\cosh(\mu s)}\right\|_{L^2_d} = \|F_{\mu,1}(s) - F_{\mu,2}(s)\|_{L^2_d} \quad (3.7)$$
$$\geq \|F_{\mu,1}(s)\|_{L^2_d} - \|F_{\mu,2}(s)\|_{L^2_d} \geq (1 - \tanh^2(\mu))\|F_{\mu,1}(s)\|_{L^2_d}.$$

Using (3.5) and $\cosh(\mu t) \geq 1$, we have

$$\|F_\mu(s)\|_{L^2_d} \geq \|\frac{F_\mu(s)}{\cosh(\mu s)}\|_{L^2_d} \geq (1 - \tanh^2(\mu))\|f(t)\|_{L^2_d}. \quad (3.8)$$

Taking $a = (1 - \tanh^2(\mu))$ yields (3.2). By (3.2), $f(t)$ must be zero if $\|F_\mu(s)\|_{L^2_d} = 0$, thus the solution of (3.1) is unique in $L^2_d(I)$. We have proven the theorem. □

**Theorem 3**. *If $f(t), F_\mu(s) \in L^2_d(I)$ and satisfy (3.1), we define the following sequence*

$$f^{(n+1)}(t) = f^{(0)}(t) + \tanh^2(\mu t) f^{(n)}(t)$$
$$- w(t) \int_{-1}^{1} \frac{K_d(t) - K_d(u)}{\pi(t-u)} \tanh(\mu u) f^{(n)}(u) du, \tag{3.9}$$

where

$$f^{(0)}(t) = \int_{-1}^{1} \frac{F_\mu(s)}{\pi(s-t)} \frac{w(t) ds}{\cosh(\mu s) w(s)}, \quad K_d(t) = \int_{-1}^{1} \frac{\tanh(\mu s)}{\pi(s-t)} \frac{ds}{w(s)}. \tag{3.10}$$

Then the sequence $\{f^{(n)}(t)\}$ converges to $\hat{f}_\mu(t) = \cosh(\mu t) f(t)$ in $L_d^2(I)$ and

$$\| f^{(n)} - \hat{f}_\mu \|_{L_d^2} \le \tanh^{2n}(\mu) \| f^{(0)} - \hat{f}_\mu \|_{L_d^2}. \tag{3.11}$$

**Proof**. Let $\hat{f}_\mu(t) = \cosh(\mu t) f(t)$, by (3.3) $\hat{f}_\mu(t)$ satisfies

$$\int_{-1}^{1} \frac{\hat{f}_\mu(t)}{\pi(s-t)} dt = \frac{F_\mu(s)}{\cosh(\mu s)} + \tanh(\mu s) \int_{-1}^{1} \frac{\hat{f}_\mu(t)}{\pi(s-t)} \tanh(\mu t) dt. \tag{3.12}$$

The left hand side of (3.12) is the FHT of $\hat{f}_\mu(t)$. By (3.6), the second term on the right hand side of (3.12) is a strictly contracting operator in $L_d^2(I)$. It then follows that the solution $\hat{f}_\mu(t)$ of (3.12) is unique in $L_d^2(I)$. Applying (1.8) to (3.12), we have

$$\hat{f}_\mu(t) = f^{(0)}(t) + \int_{-1}^{1} [\tanh(\mu s) \int_{-1}^{1} \frac{\tanh(\mu u) \hat{f}_\mu(u)}{\pi(s-u)} du] \frac{w(t)}{\pi(s-t) w(s)} ds. \tag{3.13}$$

Denote by $\hat{f}_{\mu,2}(t)$ the second term of the right hand side of (3.13). From (1.6), (1.8) annihilates the component in $F(s)$ that does not meet (1.10). It follows that $\| \hat{f}_{\mu,2}(t) \|_{L_d^2} \le \| F_{\mu,2}(s) \|_{L_d^2}$. Using (3.6), we have

$$\| \hat{f}_{\mu,2}(t) \|_{L_d^2} \le \| F_{\mu,2}(s) \|_{L_d^2} \le \tanh^2(\mu) \| \hat{f}_\mu(t) \|_{L_d^2}. \tag{3.14}$$

Define an iterative sequence $\{f^{(n)}(t)\}$ by

$$f^{(n+1)}(t) = f^{(0)}(t) + \int_{-1}^{1} [\tanh(\mu s) \int_{-1}^{1} \frac{\tanh(\mu u) f^{(n)}(u)}{\pi(s-u)} du] \frac{w(t)}{\pi(s-t) w(s)} ds. \tag{3.15}$$

Notice that $\{f^{(n)}(t)\}$ is in $L_d^2(I)$. Subtracting (3.13) from (3.15), we obtain

$$f^{(n+1)}(t) - \hat{f}_\mu(t) = \int_{-1}^{1} \{\tanh(\mu s) \int_{-1}^{1} \frac{\tanh(\mu u)[f^{(n)}(u) - \hat{f}_\mu(u)]}{\pi(s-u)} du\} \frac{w(t)}{\pi(s-t) w(s)} ds. \tag{3.16}$$

Notice that $|\tanh(\mu s)| \le |\tanh(\mu)| < 1$, thus $c = \tanh^2(\mu) < 1$. Using (3.14), we obtain

$$\| f^{(n+1)}(t) - \hat{f}_\mu(t) \|_{L_d^2} \le c \| f^{(n)}(t) - \hat{f}_\mu(t) \|_{L_d^2}, \tag{3.17}$$

and

$$\| f^{(n)}(t) - \hat{f}_\mu(t) \|_{L_d^2} \le c^n \| f^{(0)}(t) - \hat{f}_\mu(t) \|_{L_d^2}. \tag{3.18}$$

It follows that the sequence defined by (3.15) converges to $\cosh(\mu t) f(t)$ in $L_d^2(I)$. Keep in mind that the integrals of (3.15) are sequentially performed. Notice the fact that $\tanh(\mu s)/w(s) \in L_m^2(I)$ and $\tanh(\mu t) f^{(n)}(t) \in L_d^2(I)$, by Corollary 2 we simplify the second term on the right hand side of (3.15) into one-dimensional integral operator as follows

$$\int_{-1}^{1}[\int_{-1}^{1}\frac{\tanh(\mu u)f^{(n)}(u)}{\pi(s-u)}du]\frac{\tanh(\mu s)w(t)}{\pi(s-t)w(s)}ds$$

$$=\tanh^{2}(\mu t)f^{(n)}(t)-\int_{-1}^{1}\frac{w(t)}{\pi(t-u)}[\int_{-1}^{1}\frac{1}{\pi}(\frac{1}{s-t}-\frac{1}{s-u})\frac{\tanh(\mu s)}{w(s)}ds]\tanh(\mu u)f^{(n)}(u)du \qquad (3.19)$$

$$=\tanh^{2}(\mu t)f^{(n)}(t)-w(t)\int_{-1}^{1}\frac{K_{d}(t)-K_{d}(u)}{\pi(t-u)}\tanh(\mu u)f^{(n)}(u)du.$$

Here changing the order of integrals in (3.19) to a difference is based on the arguments of [14] when applying (1.5) with the pair of $f(u)=\tanh(\mu u)f^{(n)}(u)$ and $g(s)=\tanh(\mu s)/w(s)$. Thus we complete the proof. □

For $f(t),F_{\mu}(s)\in L_{m}^{2}(I)$, even for $\mu=0$, the coerciveness inequality (3.2) is not available in $L_{m}^{2}(I)$. Notice that the uniqueness holds in (1.7) if $\int_{-1}^{1}f(t)dt$ is known. Similarly, we prove that for $F_{\mu}(s)\in L_{m}^{2}(I)$, there exists a unique solution to (3.1) provided that $\int_{-1}^{1}\cosh(\mu t)f(t)dt$ is known.

**Theorem 4.** *For $F_{\mu}(s)\in L_{m}^{2}(I)$, if $\bar{f}_{\mu}=\frac{1}{2}\int_{-1}^{1}\cosh(\mu t)f(t)dt$ is known, then there exists a unique solution to (3.1) in $L_{m}^{2}(I)$. The solution can be obtained by the following sequence*

$$f^{(n+1)}(t)=f^{(0)}(t)+\tanh^{2}(\mu t)f^{(n)}(t)$$
$$-\frac{1}{w(t)}\int_{-1}^{1}\frac{K_{m}(t)-K_{m}(u)}{\pi(t-u)}\tanh(\mu u)f^{(n)}(u)du, \qquad (3.20)$$

*where*

$$f^{(0)}(t)=\frac{1}{w(t)}\int_{-1}^{1}[\frac{F_{\mu}(s)}{\cosh(\mu s)}+\bar{f}_{\mu}\int_{-1}^{1}\frac{(\tanh(\mu s)\tanh(\mu t)-1)dt}{\pi(s-t)}]\frac{w(s)ds}{\pi(s-t)}, \qquad (3.21)$$

$$K_{m}(t)=\int_{-1}^{1}\frac{\tanh(\mu s)}{\pi(s-t)}w(s)ds. \qquad (3.22)$$

*The sequence $\{f^{(n)}(t)\}$ converges to $\hat{f}_{\mu}(t)=\cosh(\mu t)f(t)-\bar{f}_{\mu}$ in $L_{m}^{2}(I)$ and*

$$\|f^{(n)}-\hat{f}_{\mu}\|_{L_{m}^{2}}\leq\tanh^{2n}(\mu)\|f^{(0)}-\hat{f}_{\mu}\|_{L_{m}^{2}}. \qquad (3.23)$$

**Proof.** If $F_{\mu}(s)\in L_{m}^{2}(I)$ and $\bar{f}_{\mu}$ is known, we rewrite (3.12) as

$$\int_{-1}^{1}\frac{\hat{f}_{\mu}(t)}{\pi(s-t)}dt$$
$$=\frac{F_{\mu}(s)}{\cosh(\mu s)}+\bar{f}_{\mu}\int_{-1}^{1}\frac{(\tanh(\mu s)\tanh(\mu t)-1)dt}{\pi(s-t)}+\tanh(\mu s)\int_{-1}^{1}\frac{\hat{f}_{\mu}(t)}{\pi(s-t)}\tanh(\mu t)dt. \qquad (3.24)$$

For simplicity, we define

$$\hat{F}_{\mu,1}(s)=\frac{F_{\mu}(s)}{\cosh(\mu s)}+\bar{f}_{\mu}\int_{-1}^{1}\frac{(\tanh(\mu s)\tanh(\mu t)-1)dt}{\pi(s-t)}. \qquad (3.25)$$

For $F_{\mu}(s)\in L_{m}^{2}(I)$, $\hat{F}_{\mu,1}(s)\in L_{m}^{2}(I)$ and (3.24) becomes

$$\int_{-1}^{1}\frac{\hat{f}_{\mu}(t)}{\pi(s-t)}dt=\hat{F}_{\mu,1}(s)+\tanh(\mu s)\int_{-1}^{1}\frac{\hat{f}_{\mu}(t)}{\pi(s-t)}\tanh(\mu t)dt. \qquad (3.26)$$

For $\hat{f}_{\mu}(t)\in L_{m}^{2}(I)$, the second term of the right hand side of (3.26) belongs to $L_{m}^{2}(I)$. Applying (2.10) to (3.26), we obtain

$$\hat{f}_\mu(t) = f^{(0)}(t) + \frac{1}{w(t)} \int_{-1}^{1} [\tanh(\mu s) \int_{-1}^{1} \frac{\tanh(\mu t) \hat{f}_\mu(u)}{\pi(s-u)} du] \frac{w(s) ds}{\pi(s-t)}. \tag{3.27}$$

Both $\hat{f}_\mu(t)$ and $f^{(0)}(t)$ are in $L_m^2(I)$. We define an iterative sequence $\{f^{(n)}(t)\}$ by

$$f^{(n+1)}(t) = f^{(0)}(t) + \frac{1}{w(t)} \int_{-1}^{1} [\tanh(\mu s) \int_{-1}^{1} \frac{\tanh(\mu t) f^{(n)}(u)}{\pi(s-u)} du] \frac{w(s) ds}{\pi(s-t)}. \tag{3.28}$$

Notice that $\{f^{(n)}(t)\}$ is in $L_m^2(I)$. Subtracting (3.27) from (3.28), we obtain

$$f^{(n+1)}(t) - \hat{f}_\mu(t) = \frac{1}{w(t)} \int_{-1}^{1} [\tanh(\mu s) \int_{-1}^{1} \frac{\tanh(\mu t)[f^{(n)}(u) - \hat{f}_\mu(u)]}{\pi(s-u)} du] \frac{w(s) ds}{\pi(s-t)}. \tag{3.29}$$

Next we analyze the property of the right hand side of (3.29). Define $\hat{F}_{\mu,2}^{(n)}(s)$ by

$$\hat{F}_{\mu,2}^{(n)}(s) = \tanh(\mu s) \int_{-1}^{1} \frac{\tanh(\mu t)[f^{(n)}(t) - \hat{f}_\mu(t)]}{\pi(s-u)} du. \tag{3.30}$$

By Lemma 2 and $c = \tanh^2(\mu) < 1$, we have $\| \hat{F}_{\mu,2}^{(n)}(s) \|_{L_m^2} \leq c \| f^{(n)}(t) - \hat{f}_\mu(t) \|_{L_m^2}$. Notice that the right hand side of (3.29) is the inverse FHT of $\hat{F}_{\mu,2}^{(n)}(s)$ by (2.10). By Corollary 1, the left hand side of (3.29) is dominated by $c \| f^{(n)}(t) - \hat{f}_\mu(t) \|_{L_m^2}$. In summary, we obtain the following estimate

$$\| f^{(n+1)}(t) - \hat{f}_\mu(t) \|_{L_m^2} \leq c \| f^{(n)}(t) - \hat{f}_\mu(t) \|_{L_m^2}, \tag{3.31}$$

and

$$\| f^{(n)}(t) - \hat{f}_\mu(t) \|_{L_m^2} \leq c^n \| f^{(0)}(t) - \hat{f}_\mu(t) \|_{L_m^2}. \tag{3.32}$$

It follows that the sequence defined by (3.28) converges to $\cosh(\mu t) f(t) - \bar{f}_\mu$ in $L_m^2(I)$. We apply the Poincaré-Bertrand formula formulated used in [14] to (3.28) with $f(u) = \tanh(\mu u) f^{(n)}(u)$ and $g(s) = \tanh(\mu s)/w(s)$, then the repeated integrals of ((3.28)) can be written into one-dimensional integral operator as follows

$$\int_{-1}^{1} [\tanh(\mu s) \int_{-1}^{1} \frac{\tanh(\mu t) f^{(n)}(u)}{\pi(s-u)} du] \frac{w(s) ds}{\pi(s-t) w(t)}$$

$$= \tanh^2(\mu t) f^{(n)}(t) - \int_{-1}^{1} \frac{1}{\pi(t-u)} [\int_{-1}^{1} \frac{1}{\pi} (\frac{1}{s-t} - \frac{1}{s-u}) \tanh(\mu s) w(s) ds] \frac{\tanh(\mu u) f^{(n)}(u)}{w(t)} du \tag{3.33}$$

$$= \tanh^2(\mu t) f^{(n)}(t) - \frac{1}{w(t)} \int_{-1}^{1} \frac{K_m(t) - K_m(u)}{\pi(t-u)} \tanh(\mu u) f^{(n)}(u) du.$$

Thus we have proven the theorem. □

**Remark 2**. When $\mu \neq 0$, the base function of null space of (3.1) is derived in [2] as follows

$$\frac{1}{\pi} \int_{-1}^{1} \frac{\cosh[\mu(s-t)]}{s-t} \frac{\cos[\mu w(t)]}{w(t)} dt = 0. \tag{3.34}$$

Due to the extra weight $\cos[\mu w(t)]$, we do not know the exact range expression of (3.1) in $L_d^2(I)$, so there may not be any solution to (3.1) for some $F(s) \in L_d^2(I)$. On the other hand, for any fixed $\bar{f}_\mu \in R^1$ in (3.24), $\{f^{(n)}(t)\}$ converges in $L_m^2(I)$, thus there are infinite number of solutions to (3.1) for every $F_\mu(s) \in L_m^2(I)$. Therefore the assumption that $\int_{-1}^{1} \cosh(\mu t) f(t) dt$ is known cannot be reduced in Theorem 4 in order for the uniqueness of the inversion.

## IV. Discretization by Cosine and Sine Transforms

Weighted spaces $L^2_d(I)$ and $L^2_m(I)$ are closely related to the Chebyshev polynomials. According to [12], $\{T_n(t), n \geq 0\}$ forms a complete orthogonal system in $L^2_d(I)$ and $\{U_n(t), n \geq 0\}$ forms a complete orthogonal system in $L^2_m(I)$. We give one definition of the Chebyshev polynomials as follows: let $U_{-1}(t) \equiv 0$, $U_0(t) = 1$, $T_0(t) = 1$, and for $n = 1, 2, ...$, define

$$T_n(t) = \cos(n \operatorname{arccos}(t)), \tag{4.1}$$

$$U_n(t) = \frac{\sin((n+1)\operatorname{arccos}(t))}{\sin(\operatorname{arccos}(t))}. \tag{4.2}$$

Here functions $T_n(t)$ and $U_n(t)$ are the Chebyshev polynomials of the first and second kind, respectively. From [22], $T_n(t)$ and $U_n(t)$ satisfy

$$\int_{-1}^{1} \frac{U_n(t)}{\pi(s-t)} \sqrt{1-t^2}\, dt = T_{n+1}(s), \tag{4.3}$$

$$\int_{-1}^{1} \frac{T_{n+1}(s)}{\pi(s-t)} \frac{1}{\sqrt{1-s^2}}\, ds = U_n(t). \tag{4.4}$$

We mention that (4.3) and (4.4) can be easily derived by using the inductive reasoning. It was mentioned in [22] that (4.3) and (4.4) may be useful to numerical computations, but we have not seen any numerical results using the Chebyshev polynomial expansion to numerically realize the FHT and its inversion. In this paper we will use the Chebyshev polynomial approximation and the discrete cosine and sine transforms to numerically evaluate the FHT and its inversion. In the remaining of this section, we always assume that $F(s) \in Ł^2_d(I)$ with the following series

$$F(s) = \sum_{n=0}^{\infty} a_n T_n(s). \tag{4.5}$$

Due to the condition of $F(s) \in Ł^2_d(I)$, $a_0 \equiv 0$ in (4.5). By (1.8), $f(t)$ can be expressed as

$$f(t) = \sqrt{1-t^2} \sum_{n=0}^{\infty} a_n U_{n-1}(t). \tag{4.6}$$

Using the coordinate transformation $\theta = \arccos(\bullet)$ from $[-1, 1]$ to $[0, \pi]$, we may express (4.5) and (4.6) in trigonometric series as follows

$$F(\cos(\theta)) = \sum_{n=0}^{\infty} a_n \cos(n\theta), \tag{4.7}$$

$$f(\cos(\theta)) = \sum_{n=0}^{\infty} a_n \sin(n\theta). \tag{4.8}$$

We call the pair of (4.7) and (4.8) the trigonometric expression of the FHT and its inversion. For a fixed integer $N$, we choose the so-called *Chebyshev-Gauss-Lobatto (CGL)* sampling points as follows

$$s_m = \cos(\frac{m+0.5}{N}\pi),\ t_m = \cos(\frac{m}{N}\pi),\ m = 0, \cdots, N-1. \tag{4.9}$$

For details on the *CGL* grid, we refer to [12, 18, 19, 25]. Define two matrices $\mathbf{C}_3 = \{c_3(m,n)\}$ and $\mathbf{S}_1 = \{s_1(m,n)\}$, $m, n = 0, \cdots, N-1$, here

$$c_3(m,n) = \sqrt{\frac{2}{N}} \begin{cases} 1/\sqrt{2} & n = 0 \\ \cos(\frac{m+0.5}{N} n\pi) & n \geq 1, \end{cases} \tag{4.10}$$

$$s_1(m,n) = \sqrt{\frac{2}{N}} \sin(\frac{m}{N} n\pi). \tag{4.11}$$

According to [12, 18, 19, 25], matrix $\mathbf{C}_3$ is the discrete cosine transform of type-III and matrix $\mathbf{S}_1$ is the discrete sine transform of type-I. As pointed out in [18], $\mathbf{C}_3$ is orthogonal and matrix $\mathbf{S}_1\mathbf{S}_1^T$ constructs a unit operator under the boundary condition of $f(t_0)=0$. On the grids of $\{s_m\}$ and $\{t_m\}$, let $\vec{a}=\{a_n\}$, $\vec{F}=\{F(s_m)\}$ and $\vec{f}=\{f(t_m)\}$, (4.7) and (4.8) can be discretized as the following linear equations

$$\vec{F} = \mathbf{C}_3\vec{a}, \quad \vec{f} = \mathbf{S}_1\vec{a}. \tag{4.12}$$

Furthermore, the discrete version of FHT and its inversion can be expressed as

$$\vec{F} = \mathbf{C}_3\mathbf{S}_1^T\vec{f}, \quad \vec{f} = \mathbf{S}_1\mathbf{C}_3^T\vec{F}. \tag{4.13}$$

We mention that (4.12) and (4.13) can be realized using the fast cosine and sine transformations if $N$ is a power of 2 [18].

Next we consider the discretization of (3.1). We will use (3.12) to discretize (3.1) and its inversion. Let $\hat{f}_\mu(t) = \cosh(\mu t)f(t)$ and $\breve{F}_\mu(s) = F_\mu(s)/\cosh(\mu s)$, we rewrite (3.12) as

$$\breve{F}_\mu(s) = \int_{-1}^{1} \frac{\hat{f}_\mu(t)}{\pi(s-t)}dt - \tanh(\mu s)\int_{-1}^{1}\frac{\hat{f}_\mu(t)}{\pi(s-t)}\tanh(\mu t)dt. \tag{4.14}$$

On $\{t_m\}$, we assume the following discrete expressions

$$\hat{f}_\mu(t_m) = \sum_{n=0}^{N-1} b_n \sin(\frac{m}{N}n\pi), \tag{4.15}$$

$$\tanh(t_m)\hat{f}_\mu(t_m) = \sum_{n=0}^{N-1} b_{1,n}\sin(\frac{m}{N}n\pi). \tag{4.16}$$

Using the trigonometric expressions (4.7) and (4.8), on $\{s_m\}$, we have

$$\breve{F}_\mu(s_m) = \sum_{n=0}^{N-1} b_n \cos(\frac{m+0.5}{N}n\pi) - \tanh(s_m)\sum_{n=0}^{N-1} b_{1,n}\cos(\frac{m+0.5}{N}n\pi). \tag{4.17}$$

The remaining task is to establish explicit formulas to derive $\{\hat{f}_\mu(t_m)\}$ and $\{\breve{F}_\mu(s_m)\}$ from each other. For $m,n=0,\cdots,N-1$, let $\delta_{m,n}$ be the Kronecker delta, we will use the following definitions:

$\vec{b} = \{b_n\}$, $\vec{b}_1 = \{b_{1,n}\}$,

$\vec{f}_\mu = \{\hat{f}_\mu(t_m)\}$, $\vec{F}_\mu = \{\breve{F}_\mu(s_m)\}$,

$\mathbf{D}_t = \{d_t(m,n)\}$ with $d_t(m,n) = \tanh(t_m)\delta_{m,n}$,

$\mathbf{D}_s = \{d_s(m,n)\}$ with $d_s(m,n) = \tanh(s_m)\delta_{m,n}$,

With such definitions, (4.15) and (4.16) become

$$\vec{f}_\mu = \mathbf{S}_1\vec{b} \quad \text{and} \quad \mathbf{D}_t\vec{f}_\mu = \mathbf{S}_1\vec{b}_1. \tag{4.18}$$

It follows that (4.17) can be expressed as

$$\vec{F}_\mu = \mathbf{C}_3\vec{b} - \mathbf{D}_s\mathbf{C}_3\vec{b}_1 = [\mathbf{C}_3\mathbf{S}_1^T - \mathbf{D}_s\mathbf{C}_3\mathbf{S}_1^T\mathbf{D}_t]\vec{f}_\mu. \tag{4.19}$$

Using (4.13), we rewrite (4.19) to the following equation

$$[\mathbf{I} - \mathbf{S}_1\mathbf{C}_3^T\mathbf{D}_s\mathbf{C}_3\mathbf{S}_1^T\mathbf{D}_t]\vec{f}_\mu = \mathbf{S}_1\mathbf{C}_3^T\vec{F}_\mu. \tag{4.20}$$

Because all the elements of $\mathbf{D}_s$ and $\mathbf{D}_t$ are dominated by $|\tanh(\mu)|<1$, the matrix on the right hand side of (4.20) is invertible and $\vec{f}_\mu$ can be obtained by

$$\vec{f}_\mu = [\mathbf{I} - \mathbf{S}_1\mathbf{C}_3^T\mathbf{D}_s\mathbf{C}_3\mathbf{S}_1^T\mathbf{D}_t]^{-1}\mathbf{S}_1\mathbf{C}_3^T\vec{F}_\mu. \tag{4.21}$$

We mention that (4.19) and (4.21) degenerate to (4.13) if $\mu=0$. The condition number of matrix $\mathbf{I} - \mathbf{S}_1\mathbf{C}_3^T\mathbf{D}_s\mathbf{C}_3\mathbf{S}_1^T\mathbf{D}_t$ has an upper bound of $[1+\tanh^2(\mu)]/[1-\tanh^2(\mu)]$. For the highest possible attenuation value $\mu=4.0$ in SPECT, the condition number is 1490.5. The accuracy of double type should be enough to avoid the ill-condition in the matrix inversion.

We cite one pair of a function and the Hilbert transform in [22] as follows

$$f_1(t) = \begin{cases} \sqrt{1-t^2} & |t| \leq 1 \\ 0 & |t| > 1, \end{cases} \qquad F_1(s) = \begin{cases} s & |s| \leq 1 \\ s - \text{sign}(s)\sqrt{s^2-1} & |s| > 1. \end{cases} \tag{4.22}$$

Through scaling and shift on (4.22), we construct the following pair

$$f(t) = \begin{cases} \sqrt{0.64-(t+0.1)^2} & -0.9 \leq t \leq 0.7 \\ 0 & \text{otherwise,} \end{cases} \tag{4.23}$$

$$F(s) = \begin{cases} s+0.1 & -0.9 \leq s \leq 0.7 \\ s+0.1 - \text{sign}(s+0.1)\sqrt{(s+0.1)^2-0.64} & \text{otherwise.} \end{cases} \tag{4.24}$$

Restricted on $[-1,1]$, $f(t)$ and $F(s)$ are continuous with boundary condition of $f(-1)=f(1)=0$. However, both $f(t)$ and $F(s)$ are not differentiable at points of -0.9 and 0.7. We will use the pair of (4.23) and (4.24) to verify the numerical properties of (4.13) on the *CGL* sampling grid $\{s_m\}$ and $\{t_m\}$ of (4.9). The number of sampling points is $N=256$ in all numerical simulations. Since $\{t_m\}$ is not evenly sampled on $[-1,1]$, for display purpose, we will resample $f(t)$ on $\{t_k\}$, here $t_k = (2k+1-N)/N$, $k=0,\cdots,N-1$. The resampling method is to first find the coefficients $\{a_n\}$ and then use the following recursive formulas

$$U_{n+1}(t_k) = 2t_k U_n(t_k) - U_{n-1}(t_k), \quad f(t_k) = \sqrt{1-t_k^2}\sum_{n=0}^{N-1} a_n U_{n-1}(t_k). \tag{4.25}$$

Similarly, we will use $\{s_k\}$, $s_k = (2k+1-N)/N$, $k=0,\cdots,N-1$, for display of $F(s)$ through the following recursive formulas

$$T_{n+1}(s_k) = 2s_k T_n(s_k) - T_{n-1}(s_k), \quad F(s_k) = \sum_{n=0}^{N-1} a_n T_n(s_k). \tag{4.26}$$

Thus, in all numerical simulations, we use the *CGL* sampling grid $\{s_m\}$ and $\{t_m\}$ for calculations but we use the evenly sampled grid $\{t_k\}$ and $\{s_k\}$ for display. We show $f(t)$ and $F(s)$ below.

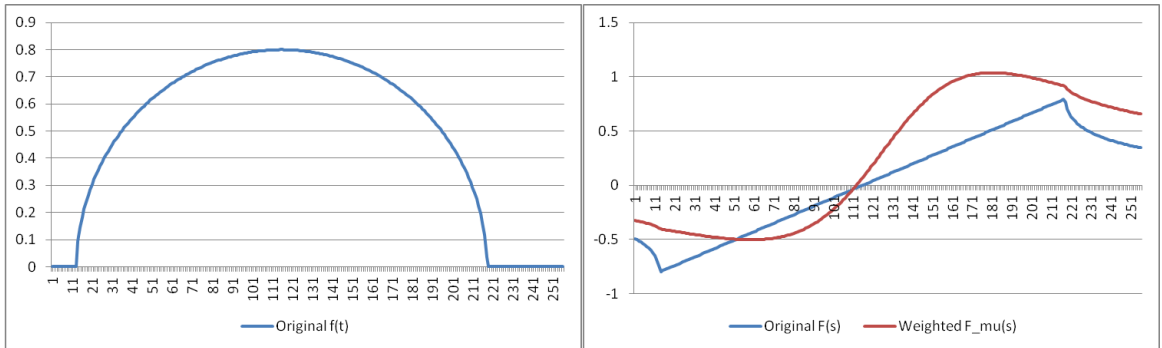

Fig 1. Left: original $f(t)$ on $\{t_k\}$; Right: $F(s)$ and $\breve{F}_\mu(s)$ on $\{s_k\}$, $\mu=3.0$.

For the analytically evaluated $\{F(s_m)\}$ by (4.24), the inverted $f(t_k)$ is shown in Fig 2.

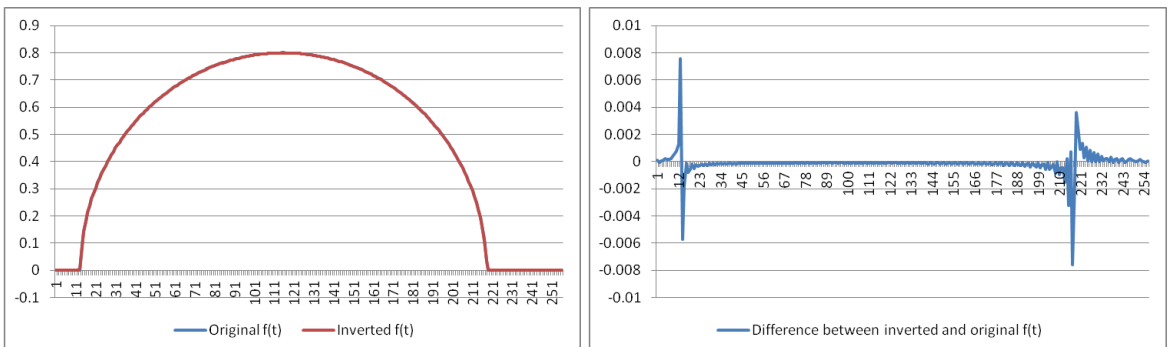

Fig 2. Left: original and inverted $f(t)$ by (4.13); Right: error of inverted $f(t)$.

We also show the numerical simulation results of (4.21) for $\mu = 3.0$. Since we do not have an explicit formula to evaluate $\breve{F}_\mu(s)$, we use a higher sampling grid with $N=512$ to numerically evaluate $\breve{F}_\mu(s)$ and then down sample it to obtain $\{\breve{F}_\mu(s_m)\}$ for $N=256$ as showed in Fig 1. The inverted $f(t)$ from such numerically generated $\{\breve{F}_\mu(s_m)\}$ is shown in Fig 3.

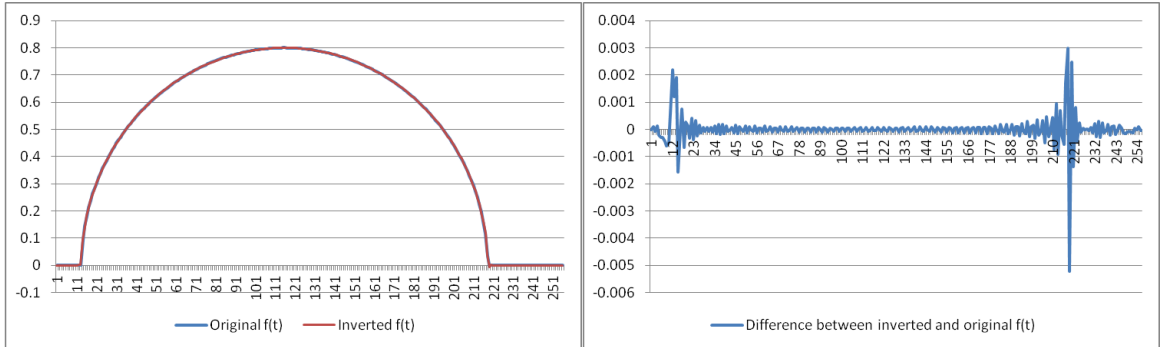

Fig 3. Left: original and inverted $f(t)$ by (4.21); Right: error of the inverted $f(t)$.

In summary, on the *CGL* sampling grid, both the FHT and its inversion can be expressed in simple trigonometric series. As a result, the numerical realization of the inverse cosh-weighted FHT becomes elementary matrix inversion (4.21).

## V.  Conclusion and Discussions

In this paper, we have obtained the Plancherel formulas and give an isotropic characterization of the FHT in $L_d^2(I)$ and $L_m^2(I)$ in Theorem 1. We have also proved the coerciveness of the cosh-weighted FHT (3.1) in $L_d^2(I)$ and the convergence of two iterative sequences (3.9) and (3.20) based on the decomposition (3.12). From theoretical point of view, we have proved the existence and uniqueness of the inversion of (3.1) in $L_d^2(I)$ and $L_m^2(I)$. This confirms that the solution from algorithms [8, 16, 21] should be unique in certain function spaces. Numerical realization of inverting (1.8) and (3.1) can be performed by a simple matrix inversion (4.21) which is much simpler than the existing numerical methods in [8, 16, 21].

So far we have only considered the inversion of (3.1) for real constant $\mu \geq 0$. As discussed in [15, 26], the constant $\mu$ may be pure imaginary numbers in reflection ultrasound tomography and seismic imaging. Let $\mu = i\eta$, then (3.1) becomes

$$F_\eta(s) = \frac{1}{\pi} \int_{-1}^{1} \frac{\cos[\eta(s-t)]}{s-t} f(t) dt . \tag{5.1}$$

For $|s| \leq 1$, $|\eta| < \pi/4$, we have $\cos(s\eta) \geq \sqrt{0.5}$ and $|\tan s\eta| < 1$. It follows that the sequences in Theorems 3 and 4 remain convergent if $|\eta| < \pi/4$. Furthermore, Theorems 3 and 4 hold for any complex $\mu$ with $|\tanh s\mu| < 1$ for $s \in [-1,1]$. For a pure imaginary $\mu$, one estimate for the stable inversion was derived in [15] using the Bessel functions in the frequency domain. In this paper we have found a different estimate for the stable inversion in the context of the FHT when only 180º projections are available in (1.11).

Now we comment on the numerical properties of (1.7) and (1.8). Formula (1.7) was used to perform the numerical experiments in [16] in which $\int_{-1}^{1} \cosh(\mu t) f(t) dt$ is required to be known and can be obtained from two projections. If two projections happen to contain larger noise or numerical errors, the individual error will propagate to the inversion globally with a factor of $1/\sqrt{1-t^2}$. Also that individual value plays an important role in the uniqueness of the inversion. In our investigation, we would rather avoid such individual datum in the inversion by (1.7) with a slightly larger interval [27, 29], and through (1.8) directly [8]. With (1.8), as shown in Theorem 1, the uniqueness of the inversion is available without extra data and the components that do not meet (1.10) will be annihilated during the inversion. Also (1.8) has explicitly trigonometric expressions in (4.12) and (4.13). For such consideration, we prefer using (1.8) in numerical simulations and indeed obtained very stable numerical results in our previous work [8] and Figs 2-3 in this paper. In particular, the fast cosine and sine transforms can be used to implement (1.8) without introducing the singularity. This is another advantage of using (1.8).

Next we mention that the algorithms in [8, 16, 21] are based on the following decomposition

$$F_\mu(s) = \int_{-1}^{1} \frac{f(t)}{\pi(s-t)} dt + \int_{-1}^{1} \frac{\cosh[\mu(s-t)]-1}{\pi(s-t)} f(t) dt . \tag{5.2}$$

The first term on the right hand side of (5.2) is the FHT and the second term defines a compact operator in $L^p(I)$. To prove the uniqueness of the inversion of (5.2), the compact operator needs to be strictly contracting. However, this has not been proved in [8, 16, 21] though the numerical results seem to suggest that the compact operator is likely to be strictly contracting. In contrast to (5.2), the decomposition (3.12) includes the FHT on the left hand side and a strictly contracting operator on the right hand side. It is this decomposition that allows us to prove the coerciveness (3.2) and the uniqueness in $L^2_d(I)$ and $L^2_m(I)$. Also the coerciveness is available in the discrete version (4.20). The big advantage of decomposition (3.12) is that the inversion of cosh-weighted FHT becomes an explicit matrix inversion (4.21).

Also we comment on the local uniqueness of the inverse FHT and ERT. A unique result was obtained in [24] if $g(x,y)$ is a piecewise polynomial in certain region of interest (ROI). The total variation method of [24] is to directly work in the subset of $R^2$ without the backprojection step (1.12). In the context of truncated Hilbert transform, we have obtained a local uniqueness result in our recent work [28] that the solution of (3.1) is unique if both $f(t)$ and $F_\mu(s)$ can be known in any small interval.

To conclude the paper, we like to mention the elegant results of [2]. Both range condition and explicit inversion formulas were obtained by using the Riemann-Hilbert problem for complex vectors. Unfortunately, we have not learned these results during the preparation of this paper, currently we are investigating the numerical characteristics of formulas in [2].